%% file: main.tex
\documentclass[11pt]{article}

\usepackage{acl}
\usepackage{microtype}
\usepackage{graphicx}
\usepackage{subcaption}
\usepackage{booktabs} 

\usepackage{hyperref}

\usepackage{amsmath}
\usepackage{amssymb}
\usepackage{mathtools}
\usepackage{amsthm}

\usepackage[table]{xcolor}
\usepackage[capitalize,noabbrev]{cleveref}

\usepackage{marvosym}

\theoremstyle{plain}

\theoremstyle{definition}

\theoremstyle{remark}

\usepackage[textsize=tiny]{todonotes}

\usepackage{multirow}
\usepackage{booktabs}
\usepackage{xcolor}

\usepackage{graphicx}
\usepackage{float}

\definecolor{lightgray}{gray}{0.95}
\definecolor{deepblue}{RGB}{70,130,180}
\definecolor{deepgray}{RGB}{119,136,153}
\definecolor{RosyBrown}{RGB}{188,143,143}
\definecolor{PeachPuff3}{RGB}{205,175,149}

\usepackage{enumitem}
\usepackage{listings}
\usepackage{xcolor} 

\lstdefinestyle{prompt}{
    basicstyle=\ttfamily\fontsize{7pt}{8pt}\selectfont,
    frame=none,
    breaklines=true,
    backgroundcolor=\color{lightgray},
    breakatwhitespace=true,
    breakindent=0pt,
    escapeinside={(*@}{@*)},
    numbers=none,
    numbersep=5pt,
    xleftmargin=5pt,
    aboveskip=2pt,
    belowskip=2pt,
}
\usepackage[most]{tcolorbox} 
\tcbset{
  aibox/.style={
    top=10pt,
    colback=white,
    enhanced,
    center,
  }
}
\newtcolorbox{AIbox}[2][]{aibox, title=#2,#1}

\usepackage{times}
\usepackage{latexsym}

\usepackage[T1]{fontenc}

\usepackage[utf8]{inputenc}

\usepackage{microtype}

\usepackage{inconsolata}

\usepackage{graphicx}

\newcommand{\modelname}{\textsc{Video-DeepResearch}}
\newcommand{\evalname}{\textsc{VideoDR-Bench}}
\title{\raisebox{-4pt}{\includegraphics[scale=0.05]{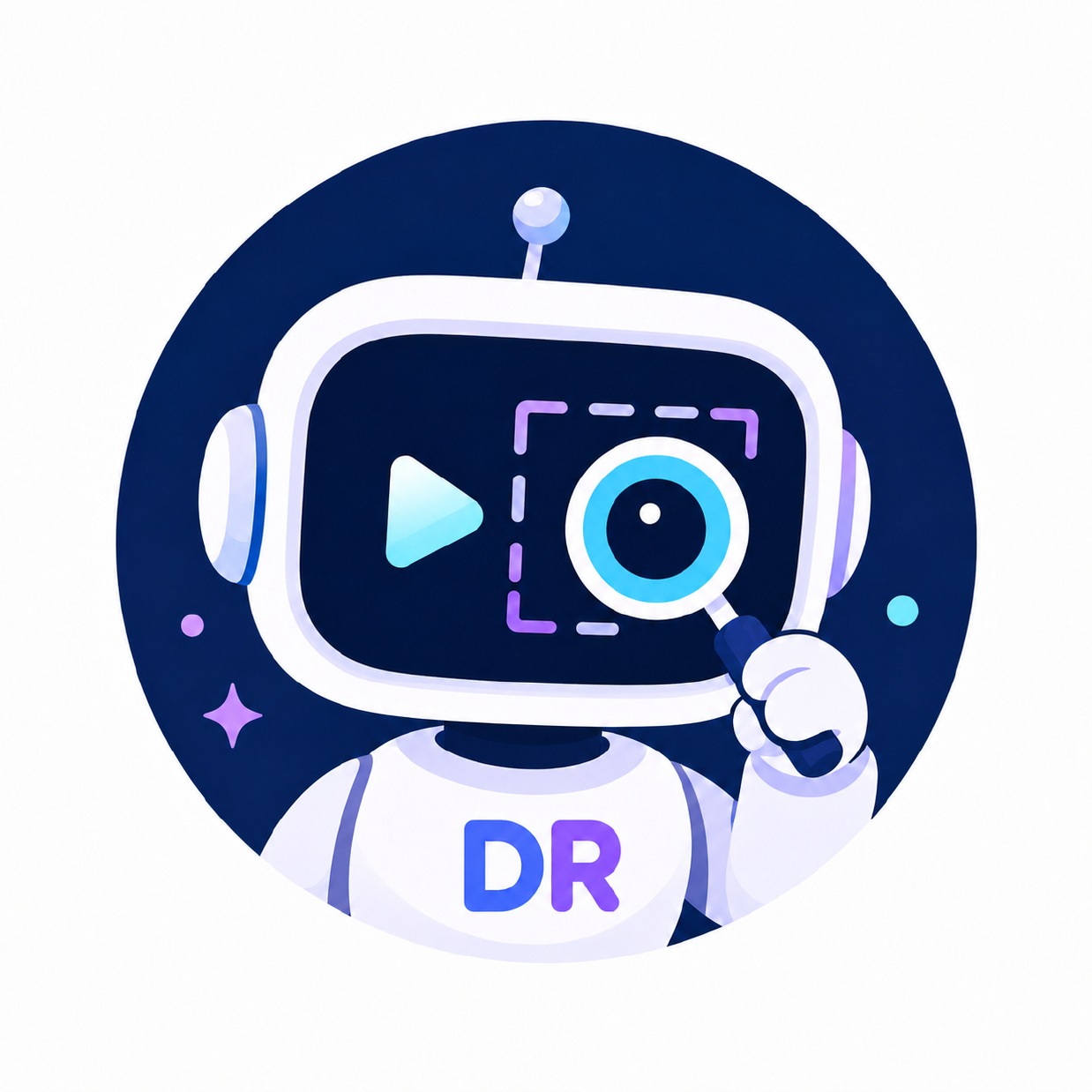}} \hspace{-1pt}Video-DeepResearch: Towards the Next-Generation Multimodal Deepresearch Agent}

\author{
\textbf{Video-Deepresearch Team} \\
\texttt{fazii@mail.ustc.edu.cn, wxhuang0616@gmail.com}\\
\texttt{https://github.com/Osilly/Vision-DeepResearch}
}

\begin{document}
\maketitle

\input{sections/0_abs}
\input{sections/1_intro}

\input{sections/3_two_issues}

\input{sections/4_method}
\input{sections/5_exp}

\input{sections/2_related}
\input{sections/6_conclusion}
\input{sections/7_impact}
\bibliography{reference}

\appendix
\input{sections/99_appendix}

\end{document}

%% file: sections/0_abs.tex
\begin{abstract}
We introduce Video-DeepResearch (Video-DR), extending multimodal agents from static images to continuous video streams, a setting that demands dense spatiotemporal grounding coupled with open-web exploration. Preliminary evaluations reveal two critical bottlenecks in current models: (1) modality bias, where agents bypass visual tools in favor of textual search, and (2) parametric knowledge leakage, where models rely on internal memory rather than genuine tool-augmented execution. To address these challenges, we propose \modelname{}, featuring a decoupled perception-exploration pipeline with stage-wise tool unlocking that compels exhaustive cross-frame visual grounding prior to web retrieval. Our framework adopts a two-stage training recipe---supervised fine-tuning followed by Group Relative Policy Optimization (GRPO)---enabling autonomous exploration that breaks the imitation-learning ceiling. Furthermore, we curate \evalname{}, a human-AI collaborative benchmark comprising 200 complex, multi-hop VQA instances. Empirical results demonstrate that our Video-DeepResearch-35B-A3B establishes a new state-of-the-art of 64.0\% average accuracy, surpassing proprietary Claude-4.5-Sonnet (59.0\%) by 5.0 points and significantly outperforming GPT-5 (52.5\%) and Gemini 2.5 Pro (57.5\%). The 30B-A3B variant achieves 59.3\%, competitive with Claude-4.5-Sonnet and demonstrating the effectiveness of our training paradigm even at compact scale.
\end{abstract}

%% file: sections/1_intro.tex
\section{Introduction}

Executing long-horizon tasks through active interaction marks a critical milestone in the pursuit of AGI, a capability epitomized by the recent rise of deep research agents~\cite{li2025websailor,wu2025webdancer,tao2025webshaper}.
Leveraging advanced VLMs~\cite{bai2025qwen2,team2026qwen3,comanici2025gemini}, research agents have advanced into unstructured, vision-rich digital wildernesses~\cite{chen2026opensearchvl,vdr,ma2025stage,mmsearch-r1,feng2026gen}. This transition shifts the agent's workload from processing clean textual inputs to decoupling critical insights from noisy, heavily redundant multimedia web layouts, redefining the complexity of autonomous exploration.

Previous literature addressing these challenges generally tracks a multi-modal trajectory. 
Text-based web agents, notably WebGPT~\cite{nakano2021webgpt} and AutoGPT~\cite{autogpt}, established the foundations of iterative knowledge synthesis through programmatic search. To accommodate rich visual layouts, subsequent paradigms shifted toward sensory interfaces; frameworks like WebWatcher~\cite{webwatcher} introduced various vision tools for better search, and recent Vision-DeepResearch~\cite{vdr} expanded this frontier to execute multi-step, exhaustive search across highly dense and complex visual contexts.

Despite these advances, current literature bypasses an ecologically valid yet far more formidable setting: \textbf{\textit{Video-DeepResearch (Video-DR)}}. This paradigm shifts the focus from isolated modalities to a holistic environment where text-based synthesis and dense visual tracking are deeply intertwined, fundamentally redefining the cognitive workload of autonomous agents.
Pioneering this Video-DR frontier entails two fundamental bottlenecks that existing methodologies fail to address.
First, on the data synthesis front, it remains largely elusive how to construct effective training pipelines that align with the intrinsic properties of video-based deep research. Conventional video datasets focus on localized captions or short-term action labels; conversely, Video-DR requires generative data curation that couples long-horizon decision trajectories with dense, time-varying textual and visual evidence.
Second, from an evaluation perspective, establishing a rigorous and high-fidelity benchmark presents a formidable challenge. Standard visual question-answering metrics are insufficient for measuring an agent's multi-step strategic execution, creating an urgent need for multi-dimensional evaluation protocols that can accurately quantify long-term reasoning consistency and error-recovery behavior under continuous temporal dynamics.

\begin{figure}[!t]
    \centering
    \includegraphics[width=\linewidth]{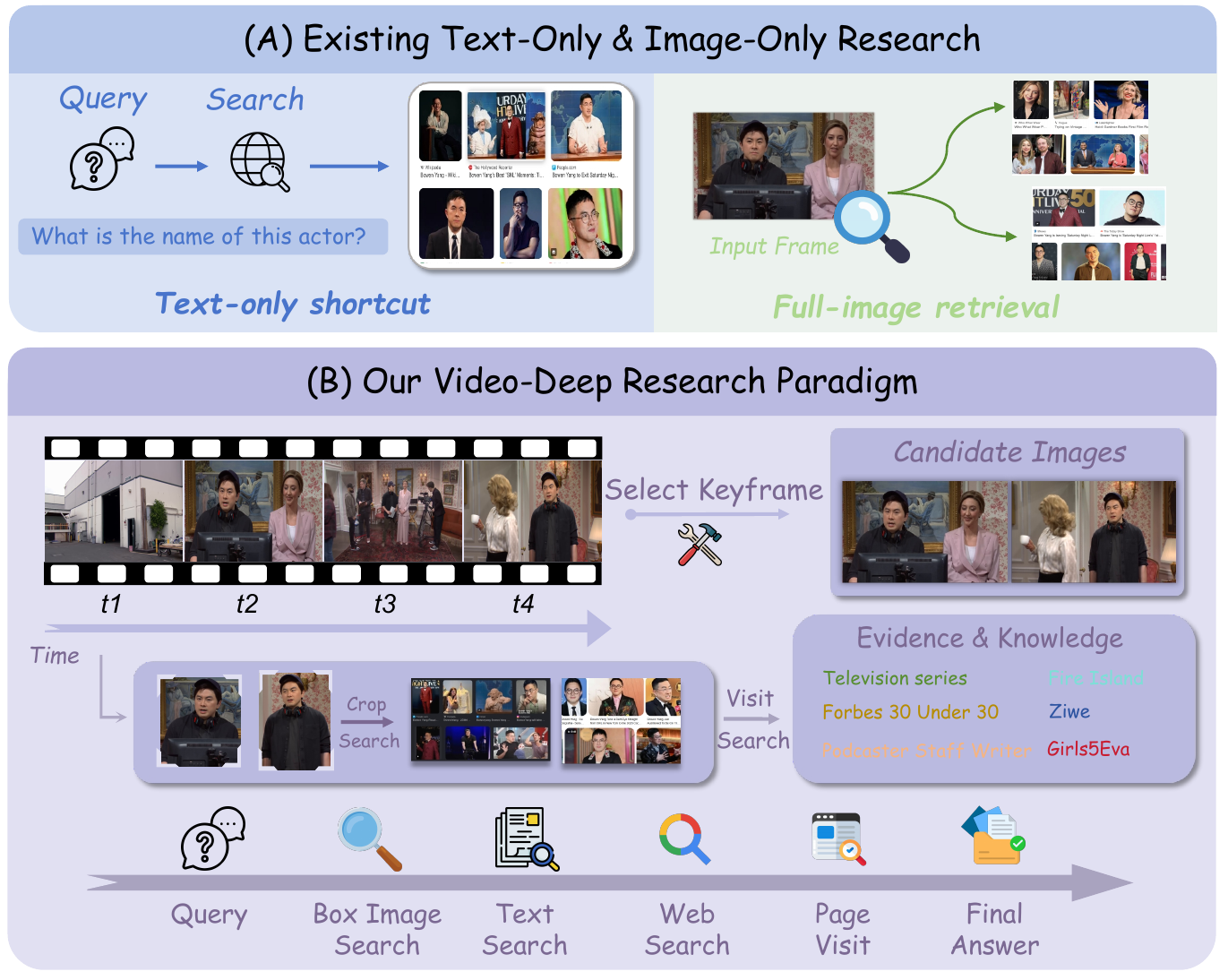}
    \caption{Overview of the {\modelname} pipeline.}
    \label{fig:pipeline}
\end{figure}
To tackle these challenges, we first conduct an empirical investigation into the fundamental failure modes of current agents when directly applied to Video-DR. By evaluating three representative models on an existing benchmark~\cite{videodr}, we uncover two striking findings: (i)~\textit{severe modality bias}, where even the strongest open-source model averages only 0.10 visual tool invocations per task while heavily relying on text search (1.27 calls), indicating a systematic aversion to active visual exploration; and (ii)~\textit{parametric knowledge leakage}, where GPT-5 attains a competitive accuracy of 57\% with virtually zero tool calls, suggesting that existing evaluations are largely solvable through memorized world knowledge alone. These findings motivate a holistic rethinking of both the training and evaluation paradigms for Video-DR agents.

Building on these insights, we propose \modelname{}, a unified framework that jointly addresses the data, training, and evaluation challenges of Video-DR. Our contributions are summarized as follows:
\begin{itemize}
    \item A scalable data engine that produces 30K video-grounded QA pairs and 7K curated trajectories through a \emph{decoupled perception-exploration} pipeline with stage-wise tool unlocking, directly countering the modality bias revealed in our preliminary study.
    \item A two-stage training recipe combining supervised fine-tuning with GRPO, enabling compact models to outperform far larger proprietary systems. Our Video-DeepResearch-35B-A3B (64.0\%) establishes a new state-of-the-art, surpassing Claude-4.5-Sonnet (59.0\%) by 5.0 points; the 30B variant (59.3\%) achieves competitive performance with the proprietary baseline while demonstrating strong open-source capability.
    \item {\evalname}, a 200-instance multi-hop VQA benchmark built via scalable human\textendash AI collaborative annotation, where every question provably requires both visual search and external knowledge reasoning.
\end{itemize}

%% file: sections/3_two_issues.tex
\section{A Naive Attempt: From Image DeepResearch to Video DeepResearch}
\label{motivation}
As shown in Fig.~\ref{fig:pipeline}, we conceptualize video as a \textbf{temporal composition of key entity trajectories}. Consequently, our key insight is that Video-DR can be formulated as a sequential grounding pipeline: identifying critical temporal frames, performing localized visual search, and synthesizing findings via subsequent text retrieval. To operationalize this workflow and bridge the gap from image-centric research, we equip the agent with two fundamental tools: \texttt{Select\_Keyframe} and \texttt{Crop\_Search}. Specifically, the agent first uses \texttt{Select\_Keyframe} to isolate informative moments from the continuous stream, and then applies \texttt{Crop\_Search} on salient entities to construct precise visual queries, which firmly anchors the downstream text-based exploration.
\begin{figure*}[!h]
    \centering
    \includegraphics[width=0.98\textwidth]{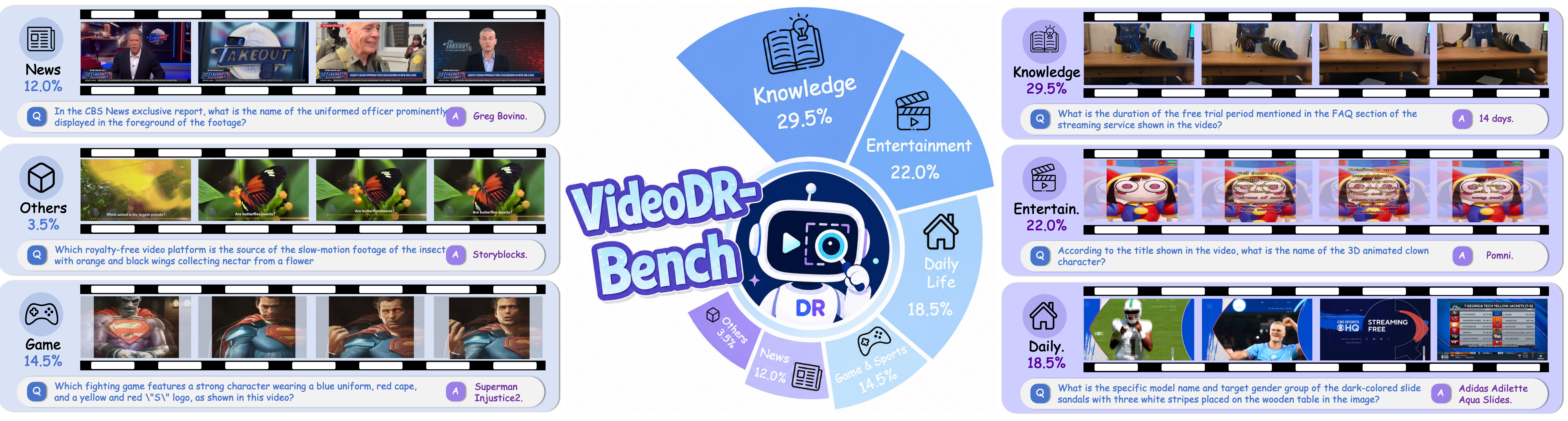}
    \caption{
Overview of {\evalname}. The benchmark spans six video domains: Knowledge (29.5\%), Entertainment (22.0\%), Daily Life (18.5\%), Game \& Sports (14.5\%), News (12.0\%), and Others (3.5\%). Every instance requires joint visual grounding and multi-hop external knowledge reasoning.
    }
    \label{fig:videohunt_overview}
\end{figure*}

\paragraph{Problem Formulation}
Formally, given a complex research query $Q$ and a visual input $V = \{v_1, v_2, \dots, v_T\}$ (which can be either a continuous video stream or a sequence of pre-sampled keyframes), the goal of a Video-DR agent is to synthesize a comprehensive response $R$. We formulate this as a sequential decision-making process. At step $i$, the agent generates an action $a_i \in \mathcal{A}$ based on the historical trajectory $\mathcal{H}_i = [Q, V, a_1, o_1, \dots, a_{i-1}, o_{i-1}]$:
\begin{equation}
    a_i \sim \pi_\theta(a \mid \mathcal{H}_i)
\end{equation}
where $\pi_\theta$ denotes the multi-modal policy and $o$ is the execution observation from the environment.
To operationalize the spatiotemporal grounding pipeline, the action space $\mathcal{A}$ encompasses our defined tools alongside standard web operations: $\mathcal{A} = \{\texttt{Select\_Keyframe}, \allowbreak \texttt{Crop\_Search}, \allowbreak \texttt{Text\_Search}, \dots\}$. 
Specifically, the temporal tool selects a specific index $t \in \{1, \dots, T\}$ to isolate an informative frame $v_t$. Subsequently, the spatial tool takes $v_t$ and a bounding box $B \in \mathbb{R}^4$ to crop a target entity, yielding a localized visual context $c_{vis} = \text{Crop}(v_t, B)$ for downstream search. Alternatively, these two tools can be encapsulated into a single joint operation. 

\paragraph{Empirical Study}
To empirically evaluate this naive formulation, we instantiate the agent with three representative models on a VideoDR~\cite{videodr} benchmark, specifically tracking the execution frequencies of each tool to analyze their behavioral patterns.
%
\input{tables/motivation}
As shown in Table~\ref{tab:motivation}, our preliminary evaluation yields two critical findings:

\noindent \textbf{Finding 1: Severe Modality Bias and Visual Tool Aversion.} Current models exhibit an inherent reluctance to invoke visual tools. Even the most capable open-source model, Qwen3.5-397B-A17B, executes an average of only 0.10 visual operations per task, overwhelmingly favoring text tools (1.27). This indicates that agents predominantly default to textual search, completely bypassing the intended active visual exploration.

\noindent \textbf{Finding 2: Susceptibility to Parametric Knowledge Leakage.} Existing evaluation setups suffer from severe prior knowledge leakage. Notably, GPT-5 achieves a highly competitive score of 57 while making virtually zero tool calls (0.00 for vision and 0.12 for text). This suggests that the model bypasses the multi-step grounding process entirely, relying solely on its vast internal memory to hallucinate or directly guess the correct answers.

%% file: tables/motivation.tex

\begin{table}[htbp]
\centering
\small
\caption{Comparison of Accuracy and Average Tool Invocation Counts between Models on VideoDR~\cite{videodr}.}
\label{tab:motivation}
\begin{tabular}{lccc}
\toprule
\textbf{Model} & \textbf{Acc.} & \textbf{Vision Tool} & \textbf{Text Tools} \\
\midrule
Qwen3.5-35B-A3B        & 41              & 0.04                            & 0.58                          \\
Qwen3.5-397B-A17B     & 58              & 0.10                          & 1.27                          \\
GPT-5     & 57              & 0.00                            & 0.12                          \\
\bottomrule
\end{tabular}
\end{table}

%% file: sections/4_method.tex
\section{{\modelname}}


\begin{figure*}[!t]
    \centering
    \includegraphics[scale=0.435]{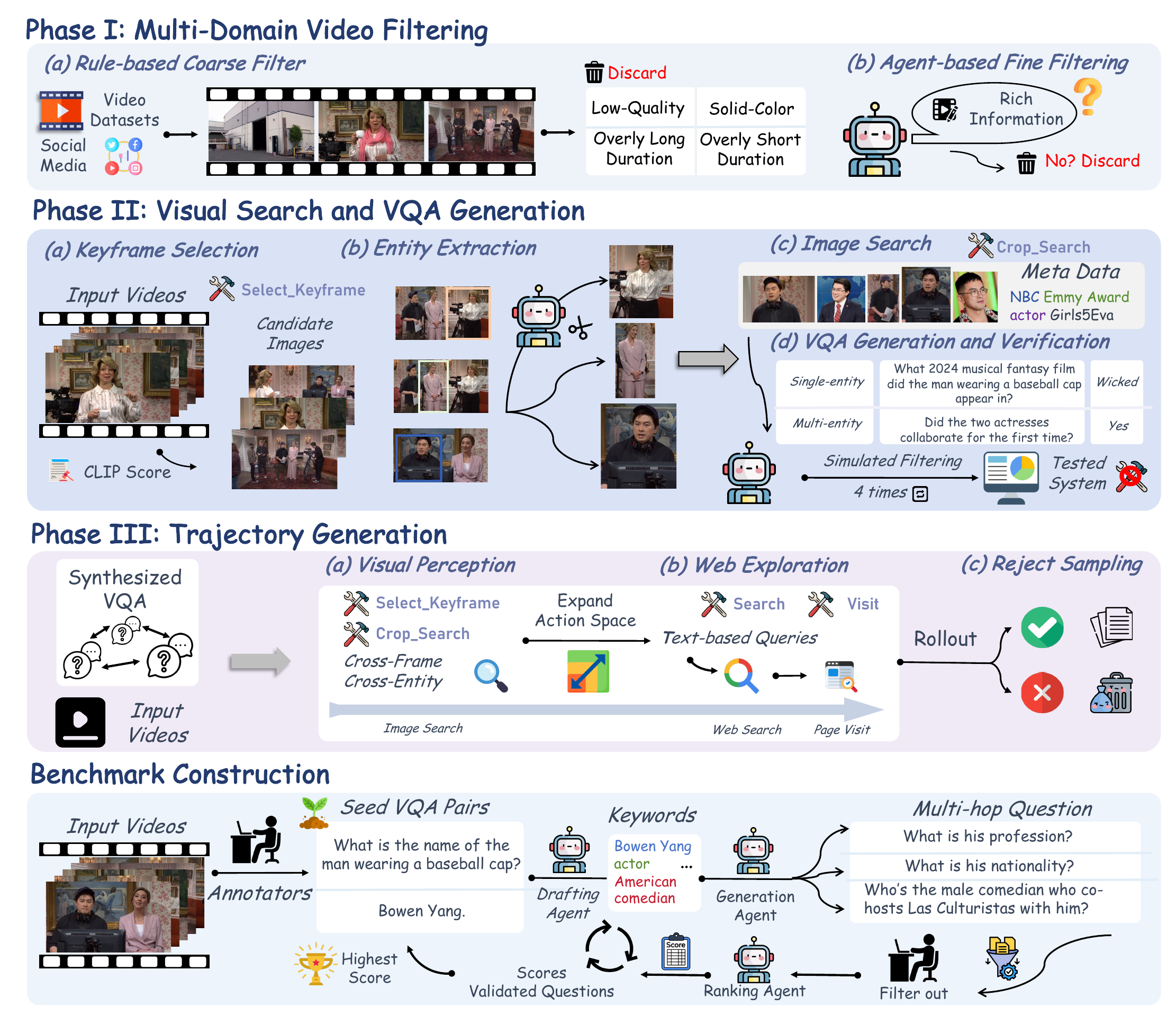}
    \caption{Overview of \textsc{Video-DeepResearch}. 
\textbf{Phase~I}: Raw videos from diverse sources are filtered via rule-based and agent-based stages. 
\textbf{Phase~II}: Keyframes are selected, entities are cropped for visual search, and VQA pairs are synthesized through single- and multi-entity patterns with parametric-leakage filtering.
\textbf{Phase~III}: Trajectories are constructed via a decoupled perception-exploration pipeline: the agent first grounds entities across frames using \texttt{Select\_Keyframe} and \texttt{Crop\_Search}, then the action space expands to \texttt{Search} and \texttt{Visit} for web exploration; only correct trajectories survive reject sampling.
    }
    \label{fig:mix}
\end{figure*}

We first detail a generative pipeline for synthesizing Video QA data (Sec.~\ref{sec:vqa}), which serves as the foundation for constructing high-quality execution trajectories (Sec.~\ref{sec:traj}). Building on this curated data, we then describe the multi-modal training procedure for our agent (Sec.~\ref{sec:train}), and finally establish a robust benchmark to evaluate Video-DR capabilities (Sec.~\ref{sec:benchmark}).

\subsection{VQA Generation}
\label{sec:vqa}
Given the absence of dedicated datasets for Video-DR, we initiate our pipeline by synthesizing foundational Video QA pairs.
This procedure transforms raw videos into explicit VQA pairs. To ensure high data quality, we annotate intermediate evidence for each instance, thereby ensuring the visual groundability and answerability of the generated queries.

\textbf{Step 0: Multi-Domain Video Filtering.}
We begin by curating a diverse collection of raw videos across multiple domains. These videos are sourced from both established video datasets~\cite{ben2025herbench, ataallah2025infinibench, wu2024longvideobench, wang2025lvbench, tao2026lvomnibench, li2024mvbench, li2025omnivideobench, goel2026mmou, li2026sekai, cheng2025video, fu2025video, hu2025video, vsibench, worldsense, yang2025longvt, wang2026videoufo} and real-world streaming platforms\footnote{YouTube}. 
These videos subsequently undergo a two-stage filtering process. First, a rule-based filter discards instances falling outside predefined duration thresholds. Second, we introduce an agentic filtering stage where Qwen3.5-35B-A3B~\cite{team2026qwen3} assesses content complexity, eliminating videos that are uninformative or overly simplistic.
Subsequently, this curated dataset is partitioned into a training set for constructing training trajectories and a test set for establishing the evaluation benchmark.

\textbf{Step 1: Keyframe Selection and Visual Search.}
We subsequently initiate an agent-driven metadata curation process. Specifically, after proposing candidate frames via CLIP-based inter-frame similarity, we deploy Qwen3.5-397B-A17B~\cite{team2026qwen3} to finalize the selection of keyframes $v_t$. For each $v_t$, the same model predicts bounding boxes $B$ to localize distinct entities $e$. These entities are then cropped and used to execute visual search queries. To ensure data fidelity, a secondary model (Qwen3.5-35B-A3B) verifies the semantic alignment between the cropped region and the retrieved results. Upon successful verification, we compile the video metadata, structured as a tuple: $\langle v_t, B, \text{entity name}, \text{search summary} \rangle$.

\textbf{Step 2: VQA Generation and Verification.}
Finally, we synthesize QA pairs from the curated metadata via two generation patterns: (1) \textit{Single-entity}: sampling one entity to formulate fact-based questions, and (2) \textit{Multi-entity}: sampling $n$ entities to construct compositional questions requiring cross-entity reasoning. In both settings, we explicitly penalize superficial visual attribute queries. Post-generation, we rigorously filter out instances prone to parametric memory leakage. Specifically, we conduct four tool-free rollouts for each question; if the agent answers correctly in any attempt, the instance is permanently discarded, guaranteeing that the remaining tasks strictly require external tool utilization.
Ultimately, we obtained 30k vqa pairs.

\subsection{Trajectory Generation}
\label{sec:traj}
Given the synthesized VQA pairs, we proceed to construct execution trajectories for agent training. As observed in Sec.~\ref{motivation}, current Video-DR agents exhibit an inherent reluctance to invoke visual tools, often bypassing them in favor of text-only search. To overcome this modality bias, we propose a decoupled trajectory construction pipeline that explicitly separates visual perception from web exploration.
Specifically, we generate trajectories using Qwen3.5-397B-A17B and apply rejection sampling. To operationalize the decoupling, we employ a stage-wise tool unlocking strategy. In the initial phase, the agent is restricted to a vision-only action space, comprising solely \texttt{Select\_Keyframe} and \texttt{Crop\_Search}. Instead of rushing to an answer, the agent is forced to execute extensive visual retrieval by cropping distinct entities across multiple keyframes. Once the agent determines the visual context is sufficient—or a predefined maximum perception horizon is reached—we expand the action space to include textual tools (\texttt{Search} and \texttt{Visit}) and prompt the agent to derive the final answer. This two-stage paradigm compels the model to conduct exhaustive cross-frame, cross-entity visual grounding prior to web exploration. Finally, we retain only the successfully resolved trajectories for downstream policy training.
Ultimately, we obtained 7k correct trajectories.

\subsection{Training}
\label{sec:train}
We adopt a two-stage training paradigm. We select Qwen3-VL-30B-A3B-Instruct~\cite{bai2025qwen2} as our base model for Video-DeepResearch-30B-A3B, given its widespread adoption as an open-source VLM and its robust AI infrastructure support. For Video-DeepResearch-35B-A3B, we adopt Qwen3.5-35B-A3B~\cite{team2026qwen3} as the foundation model, following the same training recipe. Both variants undergo identical training procedures. 
In the first stage, we perform Supervised Fine-Tuning (SFT) to establish a cold start. This phase aims to align the model with the desired decoupled perception-exploration workflow, enabling it to internalize the correct multi-modal reasoning syntax. 
In the second stage, we apply Group Relative Policy Optimization (GRPO) to further refine the policy. By actively generating rollouts and receiving rewards for successful trajectories, the agent is encouraged to autonomously explore the action space, thereby surpassing the performance ceiling of the initial SFT phase.
Experiments are conducted on a compute cluster comprising four NVIDIA H800 (80GB) GPU nodes.
More details are detailed in Appendix.~\ref{sec:train_detail}.
\paragraph{SFT}
For the SFT phase, we utilize the 7K high-quality trajectories synthesized in Sec.~\ref{sec:traj}, enabling the model to internalize the decoupled perception-exploration paradigm. Furthermore, to address the under-utilization of text tools observed in Table~\ref{tab: motivation}, we augment our training corpus with an additional 7K text-only QA instances from VDR~\cite{vdr}. This mixed-training strategy explicitly reinforces the agent's fundamental deep research capabilities. The same data recipe is applied for both the 30B and 35B variants. 
Formally, given the mixed dataset $\mathcal{D}$, where each instance consists of a context $x$ (including the system prompt, visual inputs, and interaction history) and the target output sequence $y = \{y_1, \dots, y_N\}$, the SFT objective is to minimize the standard auto-regressive negative log-likelihood:
\begin{equation}
    \mathcal{L}_{\text{SFT}} = - \mathbb{E}_{(x, y) \sim \mathcal{D}} \left[ \sum_{i=1}^{|y|} \log \pi_\theta(y_i \mid x, y_{<i}) \right]
\end{equation}
where $\pi_\theta$ represents the policy of the base model and $y_{<i}$ denotes the preceding tokens.

\paragraph{RL}
To push the agent beyond static SFT imitation and incentivize endogenous exploration, we employ Group Relative Policy Optimization (GRPO)~\cite{shao2024deepseekmath}. GRPO computes advantages via intra-group relative rewards, efficiently eliminating the memory overhead of a separate value network.
We construct a 2K moderate-difficulty RL dataset by executing four rollouts per trajectory (Sec.\ref{sec:traj}) and strictly retaining instances with a Pass@4 score between 0 and 1. We apply a sparse binary reward, assigning $r=1$ for correct answers (judged by Qwen3-VL-30B-A3B-Instruct) and $r=0$ otherwise. To prevent formatting violations or repetitive loops from dominating the updates, we compute the corresponding advantage $\hat{A}$ but down-sample their negative gradients, applying them with only a 20\% probability. The objective is defined as:
\begin{equation}
\begin{split}
    \mathcal{L}_{\text{GRPO}} = & \frac{1}{G} \sum_{i=1}^G \Big[ \min \Big( \frac{\pi_\theta(o_i)}{\pi_{\text{old}}(o_i)} \hat{A}_i, \\
    & \text{clip}\Big(\frac{\pi_\theta(o_i)}{\pi_{\text{old}}(o_i)}, 1-\epsilon, 1+\epsilon\Big) \hat{A}_i \Big) \Big] - \beta \mathbb{D}_{\text{KL}}
\end{split}
\end{equation}

\subsection{{\evalname}}
\label{sec:benchmark}
To establish our evaluation benchmark, we sample a subset from the rigorously filtered video pool. Since a robust benchmark strictly necessitates both answerability and high data fidelity, we introduce a scalable human-in-the-loop annotation framework to finalize the curation.

Specifically, given a video and an optional source URL, human annotators are instructed to pause at critical timestamps. They then utilize the \texttt{Crop\_Search} tool to query salient visual entities, strictly verifying the consistency between the retrieved external evidence and the original frame. Based on these verified results, annotators formulate several seed VQA pairs per video. Subsequently, these seeds are fed into a multi-agent framework to synthesize complex, multi-hop reasoning questions.
Specifically, the multi-agent pipeline operates as follows. First, a \textbf{Drafting Agent} brainstorms semantic directions to expand the seed VQA (e.g., expanding a base answer like ``LeBron James'' into related keywords such as his team, spouse, or MVP awards). These generated keywords are subsequently queried via a search engine. 
Next, a \textbf{QA Generation Agent} utilizes the retrieved web contexts to formulate novel multi-hop questions. To strictly enforce external tool dependency, we filter out any questions that the model can answer correctly without tool access (i.e., parametric knowledge leakage). 
Human annotators then manually verify the answerability of the remaining candidates based on the retrieved evidence. Following this, a \textbf{Ranking Agent} scores the validated questions, retaining only the highest-rated instance. Crucially, this top-ranked VQA can recursively serve as a new seed, enabling an iterative loop to synthesize increasingly complex, higher-hop reasoning tasks.
\input{tables/video_stat}

%% file: tables/video_stat.tex
\begin{table}[htbp] 
\centering
\caption{\textbf{Video Length Distribution of {\evalname}.} }
\label{tab:video_duration}
\small
\renewcommand{\arraystretch}{1.2} 
\setlength{\tabcolsep}{8pt} 
\begin{tabular}{@{} l c r r @{}}
\toprule
\textbf{Video Type} & \textbf{Duration (min)} & \textbf{Count} & \textbf{Percentage} \\
\midrule
Short       & $\le 2$       & 92 & 46.0\% \\
Medium      & $2 \sim 10$   & 68 & 34.0\% \\
Long        & $\ge 10$      & 40 & 20.0\% \\
\bottomrule
\end{tabular}
\end{table}

%% file: sections/5_exp.tex
\section{Experiments}
\subsection{Experimental Setups}
We evaluate various Vision–Language Models (VLMs), including Gemini 2.5 Pro~\cite{comanici2025gemini}, GPT-5~\cite{singh2025openai}, Claude-4.5-Sonnet~\cite{anthropic2026claude4}, Qwen3-VL-30B-A3B-Instruct~\cite{bai2025qwen2}, Qwen3.5-35B-A3B~\cite{team2026qwen3}, Qwen3.5-397B-A3B~\cite{team2026qwen3} and Kimi K2.5~\cite{team2026kimi}, on {\evalname} and VideoDR~\cite{videodr}. \evalname{} is an evaluation benchmark comprising 100 human-annotated VQA pairs, designed to assess an agent's complex reasoning capabilities by integrating video contexts with open-web exploration.

We evaluate the models under the \textbf{Agentic} setting: where the model is equipped with the full suite of visual and text tools as shown in Table~\ref{tab:tools}.
\input{tables/main_result}

Under identical interaction constraints for fairness, we extract the final prediction from the trajectory's last step. Correctness is then evaluated by Qwen3-VL-30B-A3B-Instruct, adopting the official judge prompt from Tongyi DeepResearch~\cite{team2025tongyi}.

\subsection{Main Results}
The main evaluation results are summarized in Table~\ref{tab:main_results}. Our proposed \modelname{} demonstrates exceptional deep research capabilities across multiple model scales.

\noindent\textbf{Video-DeepResearch-35B-A3B.} Establishing a new state-of-the-art among all evaluated models with an overall average accuracy of 64.0\%, our 35B variant surpasses the leading closed-source model, Claude-4.5-Sonnet (59.0\%) by 5.0 percentage points. Compared to its foundation model (Qwen3.5-35B-A3B), \modelname{}-35B demonstrates a remarkable +21.2\% improvement, with particularly strong gains across knowledge-intensive (KNL: +20.3\%) and entertainment (ENT: +15.9\%) categories. Notably, \modelname{}-35B achieves the highest score on \evalname{} (65.4\%) among all evaluated models.

\noindent\textbf{Video-DeepResearch-30B-A3B.} The 30B variant achieves 59.3\% average accuracy, competitive with Claude-4.5-Sonnet (59.0\%) and significantly eclipsing GPT-5 (52.5\%) and Gemini 2.5 Pro (57.5\%). Compared to its foundation model (Qwen3-VL-30B-A3B-Instruct), \modelname{}-30B achieves substantial absolute improvements: a +24.0\% surge on the VideoDR benchmark and a +13.5\% gain on \evalname{}. This explicit performance leap rigorously validates the efficacy of our trajectory synthesis and RL optimization pipeline even at compact scale.

\noindent\textbf{Comparison with Proprietary Models.} Despite their massive parameter counts and extensive training, proprietary models exhibit notable limitations on Video-DR tasks. GPT-5 (52.5\%) lags significantly behind, likely due to insufficient optimization for visual tool usage. Gemini 2.5 Pro (57.5\%) shows competitive performance but still falls short of both our variants. These results underscore that raw model scale alone does not guarantee effective Video-DR capability; specialized training pipelines are essential.

\noindent\textbf{Fine-Grained Category Analysis.} On the fine-grained level, our models exhibit distinct strengths across different video domains:
\begin{itemize}
    \item \textbf{Knowledge (KNL):} \modelname{}-35B achieves 66.1\%, demonstrating superior capability in factual reasoning over video-grounded knowledge queries.
    \item \textbf{Entertainment (ENT):} Both variants excel (61.4\% and 65.9\% respectively), outperforming all proprietary models.
    \item \textbf{Daily Life (DLY):} \modelname{}-35B significantly improves to 56.8\% (+16.3\% over base), indicating better generalization to common scenarios.
    \item \textbf{News (NWS):} \modelname{}-30B shows particular robustness (58.3\%), though \modelname{}-35B (41.7\%) suggests potential domain transfer challenges for the larger variant.
\end{itemize}

\noindent\textbf{Scaling Insight.} A critical observation emerges from our experiments: the improvement margins are not uniform across model sizes. While \modelname{}-30B achieves +18.8\% over its base, \modelname{}-35B achieves +21.2\%. This suggests that our training pipeline synergizes better with increased model capacity, particularly for complex multi-hop reasoning tasks. However, the relatively smaller improvement on News category (8.4\% vs. 16.6\% for 30B) indicates potential brittleness on temporally dynamic content that warrants further investigation.

\subsection{Tool Usage Analysis}
\input{tables/freq}
To understand the behavioral patterns underlying our performance gains, we profile the tool invocation frequencies across different evaluation sets in Table~\ref{tab:tool_usage}.

\noindent\textbf{Benchmark Characteristics.} From the benchmark perspective, \evalname{} rigorously compels models to execute significantly more visual and textual operations compared to VideoDR. For instance, GPT-5 increases visual tool calls from 0.00 to 0.31 and text calls from 0.12 to 1.43. This explicitly demonstrates that our benchmark circumvents parametric knowledge leakage prevalent in existing datasets, successfully enforcing genuine multi-step, tool-augmented open-web exploration.

\noindent\textbf{Methodological Analysis.} Our proposed \modelname{} exhibits a profound shift in tool utilization compared to baseline models:
\begin{itemize}
    \item \textbf{Modality Bias Correction:} Baseline Qwen3.5-397B executes only 0.10 visual operations per task while heavily favoring text tools (1.27). In contrast, \modelname{}-30B achieves 2.33 visual and 4.24 text tool invocations on VideoDR, representing a fundamental restructuring of the agent's exploration strategy.
    \item \textbf{Balanced Multimodal Search:} Driven by our carefully curated trajectory pipeline, \modelname{} internalizes active spatiotemporal perception. Augmented by mixed-text training data, this leads to a highly balanced and exhaustive multimodal search strategy that dynamically overcomes the modality bias observed in baseline models.
    \item \textbf{Scale-Appropriate Efficiency:} Notably, \modelname{}-30B achieves higher tool usage than even the 397B baseline, confirming that training methodology outweighs raw parameter count in determining agentic capability.
\end{itemize}

\noindent\textbf{Key Insight.} The strong correlation between tool usage diversity and task performance validates our core hypothesis: effective Video-DR agents must overcome the modality bias that causes models to rely on parametric knowledge. Our training pipeline successfully instills this capability, as evidenced by both the quantitative performance gains and the behavioral shift toward more exhaustive exploration.

\subsection{Ablation Study}
\input{tables/ablation}
To meticulously disentangle the contribution of each data curation and training phase, we conduct an ablation study on \modelname{}-30B, as detailed in Table~\ref{tab:abla}.

\noindent\textbf{Video-Centric SFT (\texttt{Base} $\rightarrow$ \texttt{7K-SFT}).} Starting from the baseline (40.5\%), introducing 4K synthesized Video-DR trajectories immediately yields a +5.5\% average gain. Scaling this multimodal corpus to 7K further propels the performance to 53.0\%. This explicit +12.5\% trajectory-driven surge confirms that our visual grounding data is strictly necessary for the agent to internalize the foundational perception-exploration paradigm. Interestingly, the improvement is more pronounced on \evalname{} (51.0\% vs. 43.0\% base) than on VideoDR (55.0\% vs. 38.0\% base), suggesting that visual grounding is particularly crucial for benchmark-quality evaluations.

\noindent\textbf{Text-Augmented SFT (\texttt{+7k-text-SFT}).} Subsequently, incorporating 7K text-only QA instances results in an additional +3.8\% improvement (56.8\%). This directly validates our hypothesis: explicitly injecting textual exploration data effectively mitigates the agent's initial tool-invocation bias, thereby enforcing a more comprehensive open-web search capability. The cross-modal transfer effect suggests that textual deep research skills complement and enhance visual grounding behaviors.

\noindent\textbf{RL Optimization (\texttt{+2K-RL}).} Finally, applying GRPO on the 2K moderate-difficulty dataset achieves the peak overall accuracy of 59.3\%. This final +2.5\% performance leap underscores the necessity of reinforcement learning---pushing the model beyond static imitation to execute robust, self-driven exploration trajectories. The RL phase particularly benefits \evalname{} performance (+2.0\% to 56.5\%), indicating that the learned exploration strategy generalizes well to benchmark-quality tasks.

\noindent\textbf{Cumulative Design Insight.} The incremental improvements from each training phase reveal an important principle: Video-DR capability emerges from the synergistic combination of visual grounding (trajectory SFT), textual deep research skills (mixed SFT), and autonomous exploration (RL). Neither component alone achieves optimal performance, and the full pipeline is essential for state-of-the-art results.

\subsection{Discussion: Beyond Benchmarks}
The experimental results prompt deeper reflections on the nature of video intelligence.

\noindent\textbf{Emergence Over Scaling.} Our findings challenge the prevailing assumption that sufficient model scale alone will yield emergent capabilities. The 397B parameter Qwen3.5-397B-A13B, despite its massive capacity, performs comparably to our 30B model trained with our specialized pipeline. This suggests that Video-DR capability is not merely a function of model size, but rather an emergent property that requires deliberate curriculum design---specifically, the decoupled perception-exploration paradigm we introduce. The agent must learn \emph{when} to perceive and \emph{when} to retrieve, a temporal coordination that cannot be extracted from static corpora alone.

\noindent\textbf{Active Grounding as a Test of True Understanding.} A provocative interpretation of our results concerns what tool usage actually measures. The parametric knowledge leakage observed in GPT-5 (achieving competitive accuracy with zero tool calls) reveals that benchmark performance alone can be decoupled from genuine video understanding. Our approach, by enforcing exhaustive visual grounding prior to web retrieval, effectively operationalizes a principle: \emph{understanding a video means being able to act upon it}. The shift in tool invocation patterns (from 0.10 to 2.33 visual operations) thus represents not merely behavioral modification, but a fundamental restructuring of the agent's epistemic strategy---from passive recall to active verification.

\noindent\textbf{The Parity of Modalities.} Perhaps the most counterintuitive finding is that training methodology can outweigh model scale by nearly an order of magnitude. Our 30B model achieves higher tool diversity than the 397B baseline, suggesting that modality bias is not an architectural limitation but a distributional artifact of training data. This implies that achieving true multimodal parity requires not just architectural unification, but data and training paradigm alignment---a more nuanced requirement than simply scaling model parameters.

%% file: tables/main_result.tex
\begin{table*}[ht]
\centering
\caption{\textbf{Main Results on Video Deep-Research Agent Benchmarks.} We evaluate various state-of-the-art Vision--Language Models (VLMs) under two distinct execution settings: \textbf{Direct} (tool-free baseline) and \textbf{Agentic} (equipped with the full suite of visual and text tools). Performance is reported across \textsc{Video-DR} (Acc.) and the six fine-grained categories within our proposed \evalname{} benchmark, alongside the overall average score. Column abbreviations for \evalname{} denote the corresponding video categories: \textbf{KNL} for \textit{knowledge}, \textbf{ENT} for \textit{Entertainment}, \textbf{DLY} for \textit{daily}, \textbf{G\&S} for \textit{game\&sports}, \textbf{NWS} for \textit{news}, and \textbf{OTH} for \textit{other}.}
\label{tab:main_results}
\vspace{0.5em}
\small
\setlength{\tabcolsep}{3.5pt} 
\renewcommand{\arraystretch}{1.1} 
\begin{tabular}{@{} l c ccccccc c @{}}
\toprule
\multirow{2.5}{*}{\textbf{Model}} & \textbf{Video-DR} & \multicolumn{7}{c}{\textbf{\evalname{} (\% Acc.)}} & \multirow{2.5}{*}{\textbf{Avg.}} \\
\cmidrule(lr){3-9}
 & \textbf{Acc.} & \textbf{KNL} & \textbf{ENT} & \textbf{DLY} & \textbf{G\&S} & \textbf{NWS} & \textbf{OTH} & \textbf{Overall} & \\
\midrule

\rowcolor{blue!6} 
\multicolumn{10}{c}{\textit{\textbf{Closed-Source Models}}} \\
\addlinespace[0.3em]

Gemini 2.5 Pro~\cite{comanici2025gemini} 
& 62.0 
& 54.2 
& 52.3 
& 51.4 
& 51.7 
& \underline{54.2} 
& \textbf{57.1} 
& 53.0 
& 57.5 \\
\addlinespace[0.4em]

GPT-5~\cite{singh2025openai} 
& 57.0 
& 50.8 
& 45.5 
& 48.6 
& 48.3 
& 45.8 
& \underline{42.9} 
& 48.0 
& 52.5 \\
\addlinespace[0.4em]

Claude-4.5-Sonnet~\cite{anthropic2026claude4} 
& \underline{63.0} 
& 55.9 
& 54.5 
& \underline{54.1} 
& \underline{58.6} 
& \underline{54.2} 
& \underline{42.9} 
& 55.0 
& 59.0 \\
\addlinespace[0.6em]

\rowcolor{teal!6} 
\multicolumn{10}{c}{\textit{\textbf{Open-Source Models}}} \\
\addlinespace[0.3em]

Qwen3.5-397B-A13B~\cite{team2026qwen3} 
& 58.0 
& 49.2 
& \underline{61.4} 
& 40.5 
& 55.2 
& 29.2 
& 14.3 
& 47.5 
& 52.8 \\
\addlinespace[0.4em]

Kimi K2.5~\cite{team2026kimi} 
& 61.0 
& 54.2 
& 52.3 
& 51.4 
& 51.7 
& \underline{54.2} 
& \textbf{57.1} 
& 53.0 
& 57.0 \\
\addlinespace[0.4em]

\midrule

Qwen3-VL-30B-A3B-Instruct~\cite{bai2025qwen2} 
& 38.0 
& 44.1 
& 43.2 
& 35.1 
& 51.7 
& 41.7 
& \underline{42.9} 
& 43.0 
& 40.5 \\
\addlinespace[0.4em]

Video-DeepResearch-30B-A3B (Ours) 
& 62.0 
& \underline{62.7} 
& \underline{61.4} 
& 40.5 
& \underline{58.6} 
& \textbf{58.3} 
& \underline{42.9} 
& \underline{56.5} 
& \underline{59.3} \\
\addlinespace[0.2em]

\rowcolor{green!12}
\textbf{Improvement$\uparrow$} 
& \textbf{+24.0} 
& \textbf{+18.6} 
& \textbf{+18.2} 
& \textbf{+5.4} 
& \textbf{+6.9} 
& \textbf{+16.6} 
& \textbf{0.0} 
& \textbf{+13.5} 
& \textbf{+18.8} \\
\addlinespace[0.4em]

\midrule

Qwen3.5-35B-A3B~\cite{team2026qwen3} 
& 42.0 
& 45.8 
& 50.0 
& 40.5 
& 44.8 
& 33.3 
& 28.6 
& 43.5 
& 42.8 \\
\addlinespace[0.4em]

Video-DeepResearch-35B-A3B  (Ours) 
& \textbf{68.0} 
& \textbf{66.1} 
& \textbf{65.9} 
& \textbf{56.8} 
& \textbf{62.1} 
& 41.7 
& \underline{42.9} 
& \textbf{60.0} 
& \textbf{64.0} \\
\addlinespace[0.2em]

\rowcolor{green!12}
\textbf{Improvement$\uparrow$} 
& \textbf{+26.0} 
& \textbf{+20.3} 
& \textbf{+15.9} 
& \textbf{+16.3} 
& \textbf{+17.3} 
& \textbf{+8.4} 
& \textbf{+14.3} 
& \textbf{+16.5} 
& \textbf{+21.2} \\
\addlinespace[0.2em]

\bottomrule
\end{tabular}
\end{table*}
arxiv

%% file: tables/freq.tex
\begin{table}[htbp]
\centering
\footnotesize
\caption{Average number of visual and text tool usages on VideoDR and our benchmarks.}
\label{tab:tool_usage}
\renewcommand{\arraystretch}{1.15}
\setlength{\tabcolsep}{5pt}
\begin{tabular}{lcccc}
\toprule
\multirow{2}{*}{\textbf{Model}} 
& \multicolumn{2}{c}{\textbf{VideoDR}} 
& \multicolumn{2}{c}{\textbf{VideoDR-Bench}} \\
\cmidrule(lr){2-3} \cmidrule(lr){4-5}
& \textbf{Visual} & \textbf{Text} 
& \textbf{Visual} & \textbf{Text} \\
\midrule

Claude-4.5-Sonnet 
& 1.83 & 3.38 
& 2.25 & 3.24 \\

GPT-5 
& 0.00 & 0.12 
& 0.31 & 1.43 \\

Gemini-2.5-Pro 
& 0.31 & 0.84 
& 1.93 & 2.07 \\

Qwen3.5-35B 
& 0.04 & 0.58 
& 0.20 & 0.70 \\

Qwen3.5-397B 
& 0.10 & 1.27 
& 0.04 & 2.77 \\

Base
& 1.82 & 2.21 
& 1.75 & 2.40 \\

Ours
& 2.33 & 4.24 
& 2.98 & 3.81 \\

\bottomrule
\end{tabular}
\end{table}

%% file: tables/ablation.tex
\begin{table}[htbp]
    \centering
    \small
    \caption{Performance comparison of different models on VideoDR and our benchmarks.}
    \label{tab:abla}
    \renewcommand{\arraystretch}{1.2}
    \begin{tabular}{lccc}
        \toprule
        \textbf{Setting} & \textbf{VideoDR} & \textbf{VideoDR-Bench} & \textbf{Avg} \\
        \midrule
        Base       & 38.0 & 43.0 & 40.5 \\
        4k-SFT     & 44.0 & 48.0 & 46.0 \\
        7K-SFT     & 55.0 & 51.0 & 53.0 \\
        7K-SFT+7k-text-SFT & 59.0 & 54.5 & 56.8 \\
        14k-SFT+2K-RL & 62.0 & 56.5 & 59.3 \\
        \bottomrule
    \end{tabular}
\end{table}

%% file: sections/2_related.tex
\section{Related Work}
\subsection{Video Understanding.}

Early Video-LLMs typically rely on uniform frame sampling for single-turn inference \cite{lin2024video-llava, lin2024vila, zhang2024llavavideo}. Lacking dynamic visual querying mechanisms, they are prone to error accumulation and hallucinations. While recent agentic frameworks introduce interactive tools for active fine-grained perception \cite{yang2025longvt, zhang2025thinkingvideos2508, tian2025egor1}, they remain confined to closed-world video contexts. Consequently, they struggle with knowledge-intensive tasks that require external, verifiable evidence \cite{wang2017fvqa, fu2025livevqa}. To address this, we introduce \modelname{}, a video deep research framework that couples internal video-grounding with iterative open-web exploration. By transitioning from closed-world parametric memory to an open-world collaborative verification loop, \modelname{} enables highly robust, multi-source deep reasoning.

\subsection{Multimodal-DeepResearch Systems.}
Autonomous deep research agents have rapidly evolved from text-only systems \cite{team2025tongyi,wu2025webdancer,li2025websailor,tao2025webshaper} to image-centric Vision-DR frameworks. Recent efforts employ reverse image search \cite{webwatcher}, GRPO optimization \cite{mmsearch-r1, shao2024deepseekmath}, and entity-level cropping \cite{deepmmsearch-r1} to navigate static visual contexts. However, this trajectory entirely bypasses the continuous video modality. Unlike static images, Video-DR requires agents to decouple dense, spatiotemporal dynamics and conduct multi-step verification across noisy frames, presenting a distinctly more formidable challenge.

Concurrently, establishing rigorous evaluations for this new frontier remains elusive. While current benchmarks broadly assess static multimodal factuality \cite{cheng2025simplevqa}, external knowledge grounding \cite{wang2017fvqa, chen2023can, fu2025livevqa}, and image-based search workflows \cite{mmsearch, webwatcher}, they are inherently insufficient for video streams. Moreover, the scarce efforts to establish dedicated Video-DR benchmarks suffer from a critical bottleneck: a severe reliance on labor-intensive, unscalable manual annotation. To overcome this, we introduce a highly scalable, human-AI collaborative annotation framework for robust Video-DR evaluation.

%% file: sections/6_conclusion.tex
\section{Conclusion}
We present \modelname{}, the first unified framework for Video-DeepResearch that bridges scalable data synthesis, agent training, and rigorous evaluation. A preliminary study on existing agents reveals two critical failure modes: systematic visual tool aversion and parametric knowledge leakage, both of which undermine faithful assessment of Video-DR capabilities. \modelname{} addresses these issues through a decoupled perception-exploration pipeline with stage-wise tool unlocking, producing 30K video-grounded QA pairs and 7K curated trajectories. Combined with a two-stage SFT\textendash GRPO training recipe, our Video-DeepResearch-35B-A3B achieves 64.0\% (SOTA), while the 30B-A3B variant achieves 59.3\%, competitive with Claude-4.5-Sonnet. We further introduce \evalname{}, a 200-instance multi-hop VQA benchmark where every question provably demands both visual search and external knowledge reasoning. We hope this work lays a solid foundation for next-generation video deep research agents.

%% file: sections/7_impact.tex
\section*{Author Information}

\noindent\textbf{Full Author List:}
Zhen Fang$^{1}$,
Yu Zeng$^{1}$,
Wenxuan Huang,
Yiming Zhao$^{1}$,
Shiting Huang$^{1}$,
Tianfei Ren$^{1}$,
Qi Lu$^{1}$,
Qingnan Ren$^{1}$,
Qisheng Su$^{1}$,
Lionel Z. Wang$^{4}$,
Qingyu Yin$^{5}$,
Shuang Chen$^{6}$,
Zehui Chen$^{1}$,
Lin Chen$^{1}$,
Zhenfei Yin$^{7}$,
Yao Hu$^{2}$,
Shaohui Lin$^{8}$,
Wanli Ouyang$^{3}$,
Shaosheng Cao$^{2,9}$,
Feng Zhao$^{1}$.

\noindent\textbf{Affiliations:}
$^{1}$USTC,
$^{2}$Xiaohongshu Inc.,
$^{3}$CUHK,
$^{4}$The Hong Kong Polytechnic University,
$^{5}$ZJU,
$^{6}$UCLA,
$^{7}$Oxford,
$^{8}$ECNU,
$^{9}$THU.

\noindent
The first authors with equal contribution are \textbf{Zhen Fang}, \textbf{Yu Zeng},
\textbf{Wenxuan Huang}, and \textbf{Yiming Zhao}.
Yu Zeng and Wenxuan Huang serve as the project leaders.
The corresponding authors are \textbf{Wenxuan Huang}, \textbf{Shaosheng Cao}
 and \textbf{Feng Zhao}.

\section*{Limitation}
While {\modelname} pioneers the first comprehensive pipeline integrating data construction and model training for the complex Video-DR task, this rigorous approach introduces certain trade-offs. Primarily, achieving our current level of performance incurs considerable computational overhead. To ensure high-quality data synthesis and robust model training, the framework demands substantial GPU resources, largely due to the concurrent requirements of large-scale model deployment and dynamic web search operations. Furthermore, to guarantee accurate and reliable evaluation, the construction of our benchmark currently relies on meticulous human annotation. While this ensures high fidelity of the evaluation standard, it restricts the rapid scalability of the dataset. In future work, we aim to mitigate these constraints by exploring computationally efficient pipelines, lightweight architectures, and automated LLM-based evaluation metrics to reduce human dependency.

%% file: sections/99_appendix.tex
\section{Appendix}
\label{sec:appendix}
\section{Training Details}
\label{sec:train_detail}
\textbf{Training Setup and Hyperparameters.} 
The supervised fine-tuning (SFT) was conducted on a high-performance compute cluster comprising $4$ nodes, each equipped with $8 \times 80$\,GB GPUs ($32$ GPUs in total), utilizing the Megatron-LM framework. To accommodate the ultra-long context length of $80,000$ tokens without data packing, we employed a highly optimized mixed-parallelism strategy. Specifically, we configured a Tensor Parallelism (TP) size of $4$, a Context Parallelism (CP) size of $2$, and an Expert Parallelism (EP) size of $8$. Sequence Parallelism (SP) was also enabled to further reduce the memory footprint. The training spanned $3$ epochs with a micro-batch size of $1$ and a global batch size of $64$. The learning rate was governed by a linear warmup and decay schedule, warming up over the first $5\%$ of training steps to a peak of $1 \times 10^{-5}$, and subsequently decaying to a minimum of $5 \times 10^{-7}$.

\textbf{Optimization and Efficiency Enhancements.} 
To ensure efficient training of the Mixture-of-Experts (MoE) architecture, we applied an auxiliary loss coefficient of $1 \times 10^{-6}$ for load balancing and set the expert capacity factor to $2.0$ to mitigate token dropping. Advanced MoE computational optimizations were integrated, including permute operation fusion, Grouped GEMM, and the overlapping of shared-expert computation with communication. Furthermore, strict memory optimization techniques were adopted to prevent Out-of-Memory (OOM) errors during long-context training. We utilized FlashAttention as the primary attention backend and PyTorch's expandable segments feature (\texttt{expandable\_segments:True}) to minimize memory fragmentation. Full activation checkpointing was applied uniformly at every single layer, trading computation for memory. For system-level efficiency, fused cross-entropy loss was enabled, CPU threading was optimized with $32$ OpenMP threads, and the data processing pipeline was heavily parallelized using $128$ preprocessing processes and $8$ DataLoader workers. Checkpoints were serialized in the Safetensors format every $500$ steps, deliberately omitting optimizer and random number generator (RNG) states to conserve storage overhead.

\section{Data Details}
For keyframe extraction, we compute inter-frame similarities using \texttt{CLIP-ViT-L/14@336px}~\cite{clip}. To reduce redundancy, we discard consecutive frames with a similarity score exceeding 0.8, alongside any uninformative monochromatic frames. The maximum number of keyframes per video is strictly capped at 20. Notably, this preprocessing configuration is uniformly applied across both the data synthesis pipeline and all evaluation phases, including the processing of \textsc{VideoDR}.

\section{Annotation Details}
\label{sec:annotation_detail}
The construction of \textsc{VideoHunt} involves a rigorous human annotation process to ensure benchmark quality and reliability. The entire annotation effort spans approximately three weeks.

\textbf{Annotator Team.}
We recruit a team of 8 annotators, all with prior professional experience in multimodal large language model data annotation. Each annotator possesses strong familiarity with video understanding, visual search, and multi-hop reasoning tasks.

\textbf{Quality Assurance.}
Before the formal annotation phase, all annotators undergo a structured training session covering task definitions, tool usage (e.g., \texttt{Crop\_Search}), and common pitfalls such as parametric knowledge leakage. This is followed by a qualification test on a held-out pilot set; only annotators meeting a predefined accuracy threshold are admitted to the main annotation. During production, the workflow is organized into two decoupled stages: an \emph{annotation stage}, where annotators create and verify VQA instances, and a \emph{quality inspection stage}, where a separate group of reviewers cross-checks each instance for answerability, visual groundability, and factual correctness. Instances flagged during inspection are returned for revision or discarded.

\input{prompts/filter}
\input{prompts/extract}

\input{prompts/qa_generation}
\input{prompts/verify}

\input{prompts/tarj2}

\input{tables/tools}

%% file: prompts/filter.tex
\begin{figure*}[!h]
\centering
\begin{minipage}{0.95\textwidth}
\begin{AIbox}{Prompt for \textbf{Video-DR} Task Suitability Evaluation}
\small
\textbf{[ROLE \& TASK]} \\
You are an Expert Video Content Auditor and Information Retrieval Analyst. \\
\textbf{Task:} Evaluate video content to determine if it is suitable for a Video-based Web Search (Video-DR) task. Specifically, judge whether understanding, contextualizing, or verifying the video's core content strictly requires external web search and domain-specific knowledge.

\vspace{2mm}
\hrule
\vspace{2mm}

\textbf{[PRE-CONDITIONS (Immediate Rejection)]} \\
Before detailed analysis, apply these gating rules. If any apply, immediately evaluate as \texttt{needs\_search: "No"}:
\begin{itemize}
    \item \textbf{Trivial/Everyday Events:} Videos depicting basic, universally understood actions (e.g., a person walking, generic cooking) lacking specific nuances.
    \item \textbf{Self-Contained Content:} Videos where all necessary context is explicitly provided via on-screen text, subtitles, or narration, leaving no ``information gap''.
    \item \textbf{Static/Slideshows:} Videos lacking temporal dynamics (e.g., panning across a static image or text slides).
    \item \textbf{Low Quality:} Extremely blurry, heavily occluded, or visually unintelligible videos where formulation of a meaningful search query is impossible.
\end{itemize}

\vspace{2mm}
\hrule
\vspace{2mm}

\textbf{[SUITABILITY CRITERIA (What requires a search?)]} \\
A video is ELIGIBLE (\texttt{needs\_search: "Yes"}) if it possesses high ``Semantic Depth'' and an ``Information Gap''. Look for:
\begin{itemize}
    \item \textbf{Domain-Specific Processes:} Specialized manufacturing, medical procedures, or niche industrial operations.
    \item \textbf{Specific Entities/Events:} Rare biological species, specific historical/cultural events, distinct landmarks, or specialized equipment.
    \item \textbf{Contextual Ambiguity:} Complex events demanding external knowledge to answer ``What exactly is this?'', ``Where is this happening?'', or ``Why is this happening?''.
\end{itemize}

\vspace{2mm}
\hrule
\vspace{2mm}

\textbf{[ANALYSIS WORKFLOW]}
\begin{itemize}
    \item \textbf{Extraction:} Identify the primary event, phenomenon, or subject unfolding across the video's temporal frames.
    \item \textbf{Knowledge Gap Assessment:} Ask: ``Can an average person fully comprehend this using only general knowledge?'' If Yes, external search is unnecessary.
    \item \textbf{Search Viability:} If external knowledge is required, are the visual and temporal cues distinct enough to serve as a valid visual search query?
\end{itemize}

\vspace{2mm}
\hrule
\vspace{2mm}

\textbf{[OUTPUT FORMAT]} \\
Return ONLY a valid JSON object with exact keys. No conversational text. The ``reason'' should be concise and address the semantic complexity and information gap. \\
\texttt{\{} \\
\texttt{\ \ "needs\_search": "Yes" or "No",} \\
\texttt{\ \ "reason": "<A concise, professional explanation>"} \\
\texttt{\}}
\end{AIbox}
\end{minipage}
\caption{The structured evaluation prompt for determining whether a video possesses sufficient semantic depth and information gap to require external web search for the Video-DR task usec in Sec~\ref{sec:vqa}.}
\label{fig:videodr_prompt}
\end{figure*}

%% file: prompts/extract.tex
\begin{figure*}[!h]
\centering
\begin{minipage}{0.95\textwidth}
\begin{AIbox}{Prompt for \textbf{Multi-Frame Entity Extraction} and Grounding}
\small
\textbf{[ROLE \& TASK]} \\
You are an expert visual parsing agent specialized in analyzing video keyframe sequences. \\
\textbf{Task:} Strictly follow a sequential pipeline to observe video frames, select a diverse subset of representative keyframes, and extract exactly one highly salient entity per selected frame.

\vspace{2mm}
\hrule
\vspace{2mm}

\textbf{[EXECUTION PIPELINE]}
\begin{itemize}
    \item \textbf{Step 1: Global Temporal Observation.} Carefully examine every provided frame in chronological order. Assess the visual composition (objects, individuals, logos, icons, or text) and determine the single most visually prominent and semantically salient entity within each frame.
    
    \item \textbf{Step 2: Frame Selection \& Diversification.} Based on the global observation, select a subset of \textbf{3 to 5 frames} that collectively best represent the overarching video content.
    \begin{itemize}
        \item \textit{Salience:} The dominant entity in each selected frame must be visually prominent and semantically meaningful.
        \item \textit{Diversity:} Maximize semantic diversity across the subset. Each selected frame MUST feature a \textbf{different} dominant entity to strictly avoid cross-frame redundancy.
        \item \textit{Visibility:} Prioritize frames where the target entity is clearly visible and unobstructed.
    \end{itemize}
    
    \item \textbf{Step 3: Single-Entity Grounding.} For each of the selected frames, identify and localize \textbf{exactly one} paramount entity.
    \begin{itemize}
        \item \texttt{name}: Assign a unique, descriptive noun phrase. (No duplicate names across frames).
        \item \texttt{bbox}: Provide a tight bounding box $[x_1, y_1, x_2, y_2]$ in normalized coordinates. $(x_1, y_1)$ represents the top-left corner, and $(x_2, y_2)$ represents the bottom-right corner.
    \end{itemize}
\end{itemize}

\vspace{2mm}
\hrule
\vspace{2mm}

\textbf{[STRICT CONSTRAINTS]}
\begin{itemize}
    \item Output \textbf{ONLY} a valid JSON object. No explanations, no markdown blocks, no extra keys, and no trailing commas.
    \item Bounding box coordinates must strictly satisfy: $0 \le x_1 < x_2 \le 1000$ and $0 \le y_1 < y_2 \le 1000$.
    \item Entity names \textbf{MUST} be mutually exclusive (i.e., strictly unique) across all selected frames.
\end{itemize}

\vspace{2mm}
\hrule
\vspace{2mm}

\textbf{[OUTPUT FORMAT]} \\
\texttt{\{} \\
\texttt{\ \ "selected\_frames": [} \\
\texttt{\ \ \ \ \{"frame\_index": 0, "entity": \{"name": "entity\_name", "bbox": [x1, y1, x2, y2]\}\},} \\
\texttt{\ \ \ \ \{"frame\_index": 3, "entity": \{"name": "entity\_name", "bbox": [x1, y1, x2, y2]\}\},} \\
\texttt{\ \ \ \ \{"frame\_index": 7, "entity": \{"name": "entity\_name", "bbox": [x1, y1, x2, y2]\}\}} \\
\texttt{\ \ ]} \\
\texttt{\}}
\end{AIbox}
\end{minipage}
\caption{The structured prompt for multi-frame entity extraction used in Sec.~\ref{sec:vqa}. The prompt forces the model to act as an agent that observes temporal frames, selects a diverse subset of 3 to 5 keyframes, and grounds exactly one mutually exclusive salient entity per frame.}
\label{fig:multiframe_extract_prompt}
\end{figure*}

%% file: prompts/qa_generation.tex
\begin{figure*}[!h]
\centering
\begin{minipage}{0.95\textwidth}
\begin{AIbox}{Prompt for \textbf{Knowledge-Grounded Question Generation}}
\small
\textbf{[ROLE \& TASK]} \\
You are an expert visual-language reasoning agent. You are provided with: (1) a set of localized entities extracted from a video, and (2) their corresponding retrieved external knowledge (which may reveal their ground-truth identities). \\
\textbf{Task:} Evaluate the provided multi-modal context, select the most appropriate questioning strategy, and synthesize exactly \textbf{ONE} complex, knowledge-intensive reasoning question.

\vspace{2mm}
\hrule
\vspace{2mm}

\textbf{[QUESTIONING STRATEGIES (Select Exactly ONE)]}
\begin{itemize}
    \item \textbf{Strategy 1: Single-Entity Deep Dive.} (Optimal when one entity is highly significant but lacks strong inter-entity correlation). Select ONE key entity and formulate a question that probes deeply into its specific historical, cultural, functional, or logical background (e.g., its origin, franchise, or operational mechanics).
    \item \textbf{Strategy 2: Multi-Entity Connection.} (Optimal when there is a latent historical, logical, or functional link between entities). Select AT LEAST TWO entities and formulate a question that explores the underlying relationships, shared context, or contrasts between them.
\end{itemize}

\vspace{2mm}
\hrule
\vspace{2mm}

\textbf{[STRICT GENERATION CONSTRAINTS]}
\begin{itemize}
    \item \textbf{Implicit Referencing (No Name Leaking):} Do \textbf{NOT} use or reveal any real entity names found in the external knowledge within the question. You must use natural visual referring expressions instead. \textit{(e.g., instead of naming the object, use ``the object with a red logo on the table'' or ``the structure in the background'').}
    \item \textbf{Multi-Modal Dependency:} The question must strictly demand \textbf{both} visual grounding (identifying/differentiating the target entities in the video) \textbf{and} external knowledge reasoning. It must be impossible to answer using only text (pure lookup) or only the image (pure perception).
    \item \textbf{Complexity \& Uniqueness:} The question must be specific, challenging, and necessitate multi-step reasoning. The answer should be uniquely determinable given the combined visual-textual context.
    \item \textbf{Forbidden Typologies:} Pure text-lookup queries (e.g., ``In what year was the Eiffel Tower built?'') and pure visual-perception queries (e.g., ``What color is the rabbit?'') are strictly prohibited.
\end{itemize}

\vspace{2mm}
\hrule
\vspace{2mm}

\textbf{[OUTPUT FORMAT]} \\
Return ONLY a valid JSON object. \\
\texttt{\{} \\
\texttt{\ \ "question": "A deep reasoning question based on the chosen strategy, without revealing real entity names",} \\
\texttt{\ \ "ground\_truth": "The correct answer derived from both visual context and external knowledge",} \\
\texttt{\ \ "reasoning": "Explanation of strategy selection and how answering requires BOTH visual grounding and external knowledge"} \\
\texttt{\}}

\vspace{2mm}
\textbf{Selected Entities \& Knowledge:} \\
\texttt{\{context\}}
\end{AIbox}
\end{minipage}
\caption{The structured prompt for knowledge-grounded question generation used in Sec.~\ref{sec:vqa}. The LLM acts as an agent that dynamically selects a questioning strategy and synthesizes multi-step reasoning questions while strictly adhering to visual-referencing constraints to prevent information leakage.}
\label{fig:qgen_prompt}
\end{figure*}

%% file: prompts/verify.tex
\begin{figure*}[!h]
\centering
\begin{minipage}{0.95\textwidth}
\begin{AIbox}{Prompt for \textbf{Deep Research} Answer Verification}
\small
\textbf{[ROLE \& TASK]} \\
You are an impartial judge evaluating whether a deep research report contains the correct answer. \\
\textbf{Task:} Determine if the deep research report contains the correct answer anywhere in its content.

\vspace{2mm}
\hrule
\vspace{2mm}

\textbf{[EXECUTION RULES]}
\begin{itemize}
    \item \textbf{Thoroughness:} Read through the entire research report carefully.
    \item \textbf{Search Strategy:} Look for the correct answer anywhere in the report (it may be embedded in paragraphs, tables, or sections).
    \item \textbf{Consistency Check:} Check if the information in the report is consistent with the correct answer.
    \item \textbf{Format Independence:} The answer does NOT need to be in a specific format or labeled as "final answer".
    \item \textbf{Verdict Criteria:} Answer with "yes" if the report contains the correct answer, "no" if it doesn't or contradicts it.
    \item \textbf{Reasoning:} You must provide your reasoning to justify your final verdict.
\end{itemize}

\vspace{2mm}
\hrule
\vspace{2mm}

\textbf{[INPUT DATA]} \\
\textbf{Question:} \{question\} \\
\textbf{Correct Answer:} \{reference\_answer\} \\
\textbf{Deep Research Report:} \{assistant\_answer\}

\vspace{3mm}
\textbf{[OUTPUT FORMAT]} \\
\texttt{correct: [yes/no]} \\
\texttt{reasoning: [your explanation]}
\end{AIbox}
\end{minipage}
\caption{The structured evaluation prompt for determining if a deep-research report contains the correct answer.}
\label{fig:verify_prompt}
\end{figure*}

%% file: prompts/tarj2.tex
\begin{figure*}[!h]
\centering
\begin{minipage}{0.95\textwidth}
\begin{AIbox}{Prompt for \textbf{Deep Research Agent}}
\small
\textbf{[ROLE \& TASK]} \\
You are a deep research assistant. Your core function is to conduct thorough, multi-source investigations into any topic, handling both broad open-domain inquiries and specialized academic queries. Synthesize information from credible, diverse sources to deliver a comprehensive, accurate, and objective response. When you have gathered sufficient information, enclose the definitive response within \texttt{<answer></answer>} tags.

\vspace{2mm}
\hrule
\vspace{2mm}

\textbf{[AVAILABLE TOOLS]}

\textbf{1.~\texttt{select\_crop\_search(image\_idx, bbox, goal)}} \\
Visual search over multi-frame inputs. Pick the most informative keyframes, draw a tight bounding box around the single most discriminative region (logo, jersey, scoreboard, landmark, face, on-screen text, etc.), and execute a parallel reverse-image web search per crop.
\begin{itemize}
    \item \texttt{image\_idx}: 0-indexed frame indices mapping 1-to-1 to the \texttt{image\_N} labels in the input. Must have the same length as \texttt{bbox}.
    \item \texttt{bbox}: One tight bounding box $[x_1, y_1, x_2, y_2]$ per frame, normalized to a $0$--$1000$ scale (top-left $= [0,0]$, bottom-right $= [1000,1000]$).
    \item \texttt{goal}: Two-part description: (1) the entity/object to identify, and (2) the specific fact to look up.
\end{itemize}

\textbf{2.~\texttt{search(query)}} \\
Batched web search. Supply an array of query strings; the tool retrieves top results for each query in one call.

\textbf{3.~\texttt{visit(url, goal)}} \\
Fetch one or more webpages and extract goal-conditioned information. Use after \texttt{search} to gather detailed evidence from specific pages.

\vspace{2mm}
\hrule
\vspace{2mm}

\textbf{[STRATEGY FOR VIDEO / MULTI-FRAME QUESTIONS]}
\begin{itemize}
    \item \textbf{Vision-First:} When the input contains video frames, the \textbf{first} tool call \textbf{must} be \texttt{select\_crop\_search}. Do NOT call \texttt{search} before obtaining a concrete visual entity.
    \item \textbf{Frame Selection:} Skim every frame, then pick the 1--3 frames where the target entity is most clearly visible. Avoid blurry, occluded, or redundant frames.
    \item \textbf{Tight Cropping:} Draw a tight bounding box around the most discriminative region. Avoid loose full-frame boxes, as they retrieve noise.
    \item \textbf{Precise Goals:} State exactly what to identify (name, title, who/what/when), not vague phrases like ``search this image''.
    \item \textbf{Verify \& Expand:} Only after \texttt{select\_crop\_search} returns concrete entities should you call \texttt{search}\,/\,\texttt{visit} to verify or expand.
\end{itemize}

\vspace{2mm}
\hrule
\vspace{2mm}

\textbf{[EXAMPLE]} \\
\textit{Question:} ``Which match is shown? Give tournament, stage, teams and score.'' Frame 1 has the scoreboard at the top: \\[2pt]
\texttt{<tool\_call>} \\
\texttt{\{"name": "select\_crop\_search", "arguments":} \\
\texttt{~~\{"image\_idx": [1], "bbox": [[10, 38, 990, 200]],} \\
\texttt{~~~"goal": "Identify the football match shown by this} \\
\texttt{~~~scoreboard -- extract tournament name, stage,} \\
\texttt{~~~the two teams, and the score."\}\}} \\
\texttt{</tool\_call>}

\vspace{2mm}
\hrule
\vspace{2mm}

\textbf{[OUTPUT FORMAT]} \\
Return each function call as a JSON object within XML tags: \\
\texttt{<tool\_call>} \\
\texttt{\{"name": "<function\_name>", "arguments": \{...\}\}} \\
\texttt{</tool\_call>} \\[2pt]
When ready to finalize, enclose the complete answer within: \\
\texttt{<answer>}~\textit{your final response}~\texttt{</answer>}
\end{AIbox}
\end{minipage}
\caption{The prompt for the Deep Research Agent, which has access to all three tools and follows a vision-first strategy: grounding visual entities via \texttt{select\_crop\_search} before textual web exploration.}
\label{fig:deep_research_agent_prompt}
\end{figure*}

%% file: tables/tools.tex
 \begin{table*}[ht]
    \centering
    \small
    \renewcommand{\arraystretch}{1.25}
    \caption{Tools available to the agent. \texttt{select\_crop\_search} is the
             only tool exposed during the exploration phase; \texttt{search} and
             \texttt{visit} are added in the answering phase.}
    \label{tab:tools}
    \begin{tabular}{p{3.4cm} p{3.0cm} c p{6.3cm}}
      \toprule
      Parameter & Type & Req. & Description \\
      \midrule

      \multicolumn{4}{l}{\textbf{\texttt{select\_crop\_search}} — pick frames,
        crop a bounding box from each, run a reverse-image / visual web search
        per crop.} \\
      \addlinespace[2pt]
      \texttt{selections}
        & list of objects (1--8)
        & \checkmark
        & A batch of (frame, bbox) selections to crop and search in a single
          call. Use several selections to cast a wide net, or a single tight
          crop to zoom in on one discriminative region. \\
      \quad\texttt{.frame\_index}
        & integer ($\geq 0$)
        & \checkmark
        & 0-based index of the frame in the video's frame list. \\
      \quad\texttt{.bbox}
        & list[float], length 4
        & \checkmark
        & Crop region in normalised $[x_1, y_1, x_2, y_2]$ coordinates with
          $0.0 \le x_1 < x_2 \le 1.0$ and $0.0 \le y_1 < y_2 \le 1.0$.
          Use $[0, 0, 1, 1]$ for the full frame. \\
      \texttt{goal}
        & string
        & \checkmark
        & Precise statement of what to identify \emph{and} the fact to look up
          about it. \\
      \multicolumn{4}{p{15.5cm}}{\emph{Returns:} for each crop, the top web
        results (title, URL, snippet) and any recognised entities (people,
        products, logos, landmarks).} \\
      \midrule

      \multicolumn{4}{l}{\textbf{\texttt{search}} — run a web search and return
        the top results from Google / Serper / Zhipu (with automatic fallback).} \\
      \addlinespace[2pt]
      \texttt{query}
        & string \emph{or} list[string]
        & \checkmark
        & A single search query or a list of queries run in parallel. Prefer
          several short, focused queries over one long compound query. \\
      \texttt{num\_results}
        & integer
        & ---
        & Max results returned per query (default \texttt{10}). \\
      \multicolumn{4}{p{15.5cm}}{\emph{Returns:} per query, an ordered list of
        $\{\,\texttt{title},\,\texttt{url},\,\texttt{snippet}\,\}$ entries.} \\
      \midrule

      \multicolumn{4}{l}{\textbf{\texttt{visit}} — fetch one or more web pages
        and have a summary model distil them into a goal-conditioned JSON.} \\
      \addlinespace[2pt]
      \texttt{url}
        & string \emph{or} list[string]
        & \checkmark
        & The URL to fetch, or a list of URLs to fetch in parallel. Typically a
          URL returned by an earlier \texttt{search} call. \\
      \texttt{goal}
        & string
        & \checkmark
        & Precise statement of what to look for on the page; the summary model
          uses it to decide what to keep vs.\ discard. \\
      \multicolumn{4}{p{15.5cm}}{\emph{Returns:} per URL, a JSON object with
        \texttt{rational} (why the page is relevant), \texttt{evidence}
        (verbatim supporting spans), and \texttt{summary} (one-paragraph
        synthesis). Raw HTML is not exposed to the agent.} \\

      \bottomrule
    \end{tabular}
  \end{table*}